\documentclass{article1}
\usepackage{planrobo}
\usepackage[utf8]{inputenc}
\usepackage{times}
\usepackage[numbers]{natbib}
\usepackage{multicol}
\usepackage{hyperref}
\usepackage{graphicx}
\usepackage{epspdfconversion}
\usepackage{subcaption}

\title{Planning Robot Motion using Deep Visual Prediction}

\author{Meenakshi Sarkar, Prabhu Pradhan and
Debasish Ghose\\ GCDSL, Department of Aerospace Engineering, \\ Indian Institute of Science (IISc), \\ Bangalore, KA-560012, India\\{\tt\small (meenakshisar, dghose)@iisc.ac.in, prabhupradhan@pm.me}}

\begin{document}
\maketitle
\begin{abstract}
 In this paper, we introduce a novel framework that can learn to make visual predictions about the motion of a robotic agent from raw video frames. Our proposed motion prediction network (PROM-Net) can learn in a completely unsupervised manner and efficiently predict up to 10 frames in the future. Moreover, unlike any other motion prediction models, it is lightweight and once trained it can be easily implemented on mobile platforms that have very limited computing capabilities. We have created a new robotic data set comprising LEGO Mindstorms moving along various trajectories in three different environments under different lighting conditions for testing and training the network. Finally, we introduce a framework that would use the predicted frames from the network as an input to a model predictive controller for motion planning in unknown dynamic environments with moving obstacles. 
 \end{abstract}
\section{Introduction}

Prediction is often considered as one of the key and fundamental components of intelligence \cite{bubic}. Visual prediction often deciphers much useful information about the environment in an information-rich but high dimensional format which presents both opportunities and challenges \cite{ebert}. A successful mechanism capable of predicting future video frames would have many applications in the automation industry. For robotics, the path planning problem in an unknown dynamic environment with moving obstacles still largely remains an unsolved challenge. A key component of the problem that remains unsolved is to make predictions about the motion of the obstacles. To illustrate the complexity of the problem, consider an Unmanned Aerial Vehicle (UAV) flying at a speed of 5m/s or higher (which translates to a speed of $\geq$ 18km/hr) in a cluttered environment, such as a forest trail in a high wind condition. In this scenario, the perception of the camera not only depends upon the dynamics of the objects present in the scene but also on the control actions taken by the UAV. The interaction between the robot's state and action with the dynamics of the scene renders the motion prediction problem almost impossible to solve using conventional vision-based methods such as visual servoing \cite{visp}, \cite{vkumar}. While making pixel level prediction on the motion on robotic agents is a challenging task, designing a motion planner based on raw predicted image frames is an even harder task. However, it is imperative to have a mechanism to predict the motion of the other objects present in the environment in order to solve the motion planning problem for autonomous agents operating in a rapidly changing environment.

Recent works (\cite{villegas}, \cite{xu}) on motion prediction delved into forecasting human motion but these models use very deep architectures that ultimately renders them computationally expensive. Given that human motion is much slower than automated vehicles, predicting higher speed motion  will have a much higher level of difficulty and computation cost. Even with the recent advancements in mobile graphics units such as NVIDIA Jetson boards, implementation of deep architectures to solve path planning problems on small mobile agents still remains a challenge. 

Designing a light-weight motion prediction framework is only the first step in addressing the challenge. Once the prediction network is designed we need to devise a mechanism to transfer raw predicted image frames into control commands for the robot. Recent advancements in deep reinforcement learning (DRL) \cite{mnih}, \cite{levine} have shown us ways to convert raw sensory inputs into meaningful control commands. While a few model-free learning algorithms have out-performed human operators \cite{mnih}, these frameworks were designed for very limited simulated environments of video games. Learning tasks in the real world present a wide range of challenges as the environment becomes dynamic with sparsely available reward feedback while the agent can only access a partial state of the world. Finally, we need to design the entire framework in such a manner that learning can be enabled without the supervision of human operators.

In this work, we present a novel light-weight framework that can forecast the trajectory of an object moving in the robot's work-space. The proposed Predicting Robot Motion Network (PROM-Net)  can easily be trained on raw video data without supervision. Once trained, this network can be implemented on an autonomous mobile agent. The network generates the visual prediction of the surrounding environment from the first-person perspective of the robotic agent. In order to train and test the network we also created our own data set, using two LEGO Mindstorms under 4 different scenarios. To the best of our knowledge, this is the first data set of its kind where the motion of a robotic agent is captured from the first-person view of another robot. We also discuss how these predicted frames can be used to design a model-based reinforcement learning algorithm that would be able to translate the raw predicted image frames into a meaningful reward function to optimize the trajectories of the control policies.

The paper is organized as follows: We first discuss the existing literature on video prediction networks and model predictive controllers (MPC) using raw image frames as input and introduce the Predicting Robot Motion Network (PROM-Net) model. Then we discuss the virtual experimental setup we created in the OpenAI-Gym framework and give a detailed description of the real robotic dataset that was created for testing the performance of PROM-Net. A detailed analysis of the performance of PROM-Net is presented next, followed by a discussion on the future scope of the work.

\section{Related Work}
The problem of video frame prediction \cite{mathieu}, \cite{vondrick2},  \cite{villegas}, \cite{xu} has gained considerable popularity in the computer vision community in recent years. However, video data comes with the issue of larger dimensionality with complex spatio-temporal dynamics in raw pixel values which makes the pixel label frame prediction task very challenging \cite{finn}. While Convolutional Neural Networks (CNN) have proven to be very successful at learning features from static images,  \cite{imagenet}, \cite{resnet}, the idea of Convolutional Long Short Term Memory (LSTM) networks that were designed specifically to capture the spatial and temporal dependencies in video data was proposed by \cite{convlstm}.
\begin{figure}
\centering
  \includegraphics[width=\linewidth]{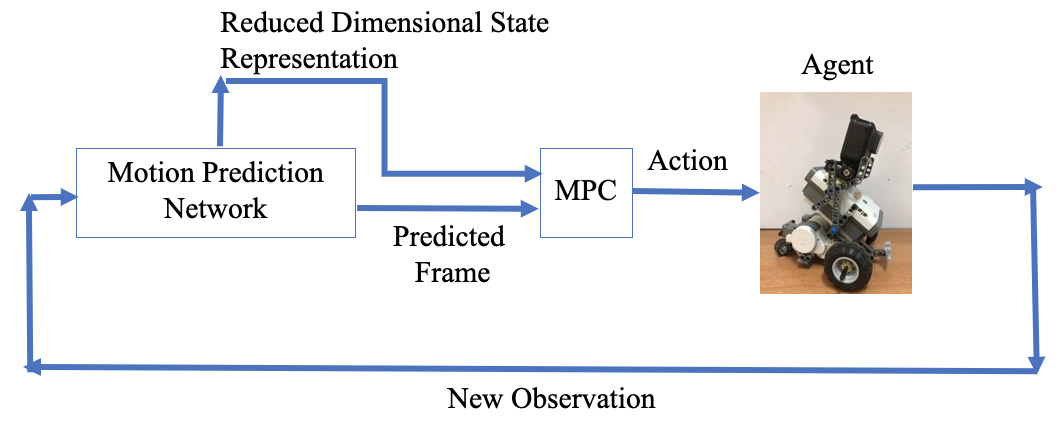}
  \caption{ Visual motion planning framework}
  \label{ICAPS_fig1}
\end{figure}
The paper \cite{oh}, designed an action conditional encoder-decoder network predicting future frames for Atari games. The work in \cite{mathieu} employed a new adversarial loss function for additional regularization and sharper frame prediction. The paper \cite{vondrick2} designed a multi-scale feedforward architecture combined with an adversarial objective to generate a foreground-background mask to create realistic looking video sequences. The work in \cite{casas} presented a framework that predicts the intention of autonomous cars from 3D point clouds and HD maps. The paper \cite{walker} proposed a framework that generated a coarse hallucination of the possible future events from a wide angle view. In \cite{xu}, a framework that balances between reconstruction and adversarial loss for the predicted frames is designed. However, most of the current state of the art video prediction models often require a high-end GPU enabled system to train and test the networks which is not often a feasible option for robotic applications.
 
  \begin{figure*}[h]
	\centering
	\includegraphics[width=.98\textwidth, height=4cm]{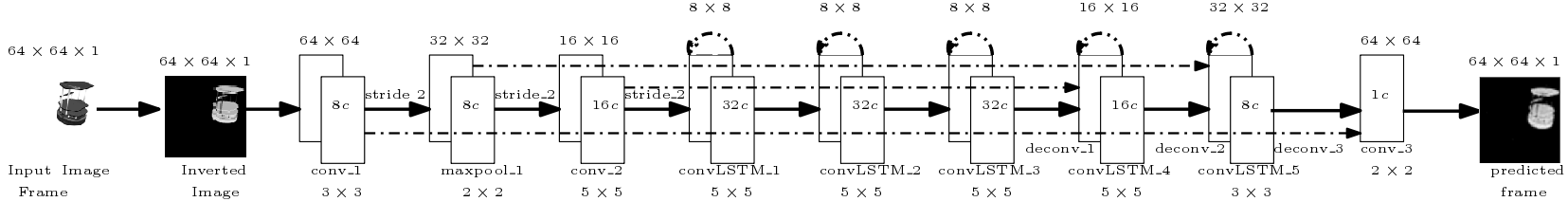}
	\caption{Schematic architecture of the PROM- Network}
	\label{ICAPS_fig2}
\end{figure*}

 While considerable progress has been made in DRL \cite{mnih},\cite{mnih2} that learns meaningful skills directly from high dimensional raw sensory data (especially images), most of these are restricted to simulated applications of computer games. Only a few works like \cite{finn2}, \cite{ebert} talks about the application of a model based RL algorithm for robotic manipulation tasks using visual foresight. To the best of our knowledge, there is no existing work that addresses the problem of an end to end motion planning for autonomous mobile agents using visual prediction from a first person (robot) perspective.

 \section{Our Approach}
 Figure \ref{ICAPS_fig1} shows the schematic representation of the visual prediction based motion planning framework. It shows that the MPC algorithm takes the frames generated by the prediction network as input. The prediction network also generates a reduced dimensional state representation of the world from the raw image inputs for the model-based controller. This is possible as the architecture of the prediction network is based on the encoder-decoder network philosophy.  We started with designing the prediction network with a goal of predicting the next $N$ image frames from the past $N$ number of frames.  Furthermore, we aim to design a very light-weight network that can be easily implemented on a GPU denied environment. We have successfully designed a motion prediction network that can approximate frames up to 10 time stamps ahead of time. In the following sub-section we present a detailed description of our proposed network.

 \subsection{Predicting Robot Motion Network (PROM-Net)}
The architecture of the model is shown in figure \ref{ICAPS_fig2}. This model roughly follows the encoder-decoder philosophy of autoencoder networks. The encoder network is built using 8 2D convolutional filters of size $3\times 3$. The outputs are down-sampled using a maxpooling layer of stride 2. A second 16 channel convolution layer with filter size $5\times5$ and stride 2, further maps the input to a 3 dimensional tensor of size $(16\times16\times16)$. These spatial feature tensors are then passed through two consecutive Convolutional LSTM layer having kernel size of $(5\times5)$ and mapped into a 32 channel feature space of size $(8\times 8)$. The mathematical model of Convolutional LSTM is described in \cite{convlstm}. The two ConvLSTM layers capture the spatio-temporal correlations present in the sequence of image frames and pass them to the next decoder layer for inference. 

The decoding network consists of 3 Convolutional LSTM layers. After each ConvLSTM layer we have a deconvolution or transpose convolution layer that upsamples the size of each feature channel and downsamples the total number of feature channels. For example, after the first deconvolution operation, the $(32\times8\times8)$ feature tensor is mapped into a $(16\times16\times16)$ feature tensor. We have used skip connections at intermediate layers to recover from the lossy convolution operations (shown with dotted lines in fig. \ref{ICAPS_fig2}).

We apply batch normalization operation after each Convolutional LSTM layer. We also upsample the number of feature channels each time we apply a downsampling operation on the 2 dimensional spatial feature space. This kind of convention has been followed in designing various previous networks such as \cite{unet},\cite{finn}. All the convolutional filters use the ReLU activation function. The entire network is trained using the RMSProp algorithm that minimizes the mean square loss.

PROM-Net has about 6 million trainable  parameters and once trained the network weighs only about 5 Megabytes.

\begin{figure*}
        \centering
        \begin{subfigure}{.5\textwidth}
        \centering
        \includegraphics[width=.8\textwidth]{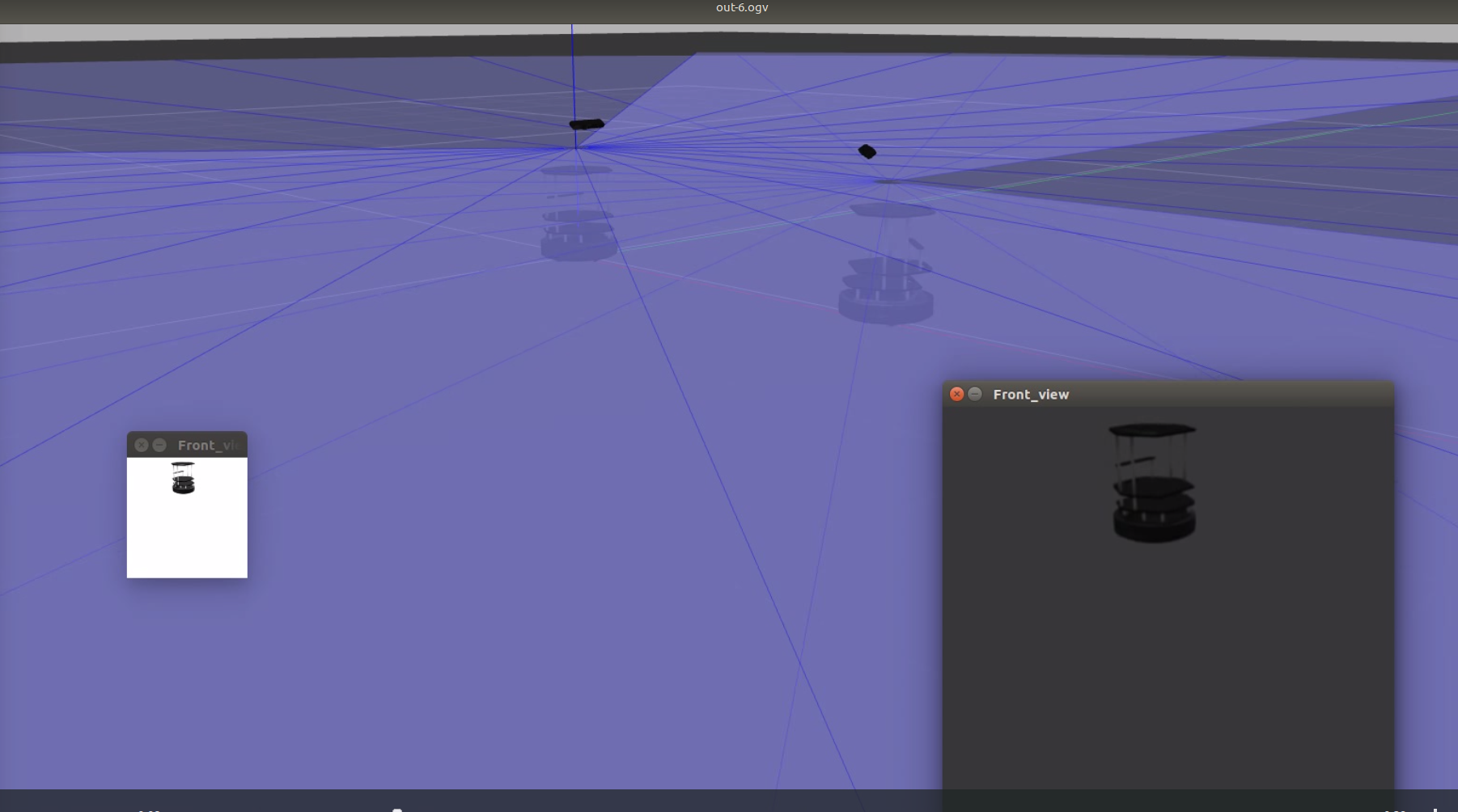}
       
        \caption{A ROS-Gazebo based virtual experimental environment has been set up in the OpenAI-gym framework.}
        \label{ICAPS_fig3a}
    \end{subfigure}%
    \hspace{1in}
	\begin{subfigure}{.3\textwidth}
  \centering 
  \includegraphics[width=0.75\linewidth,keepaspectratio]{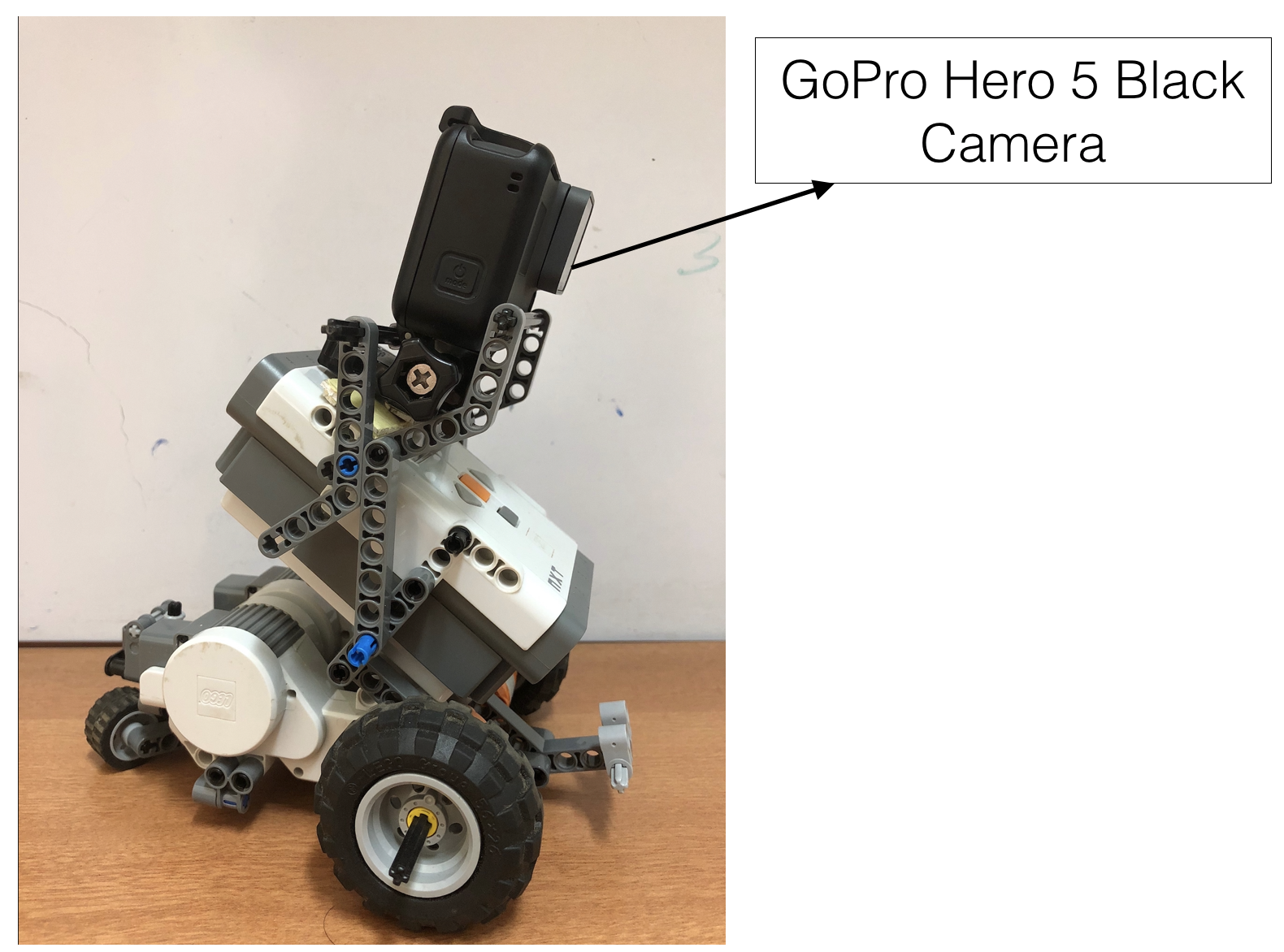}
	\caption{LEGO Mindstorms with a \\ GoPro Hero 5 Black camera}
	\label{ICAPS_fig3b}
\end{subfigure}
\caption{}
\end{figure*}%

\section{Virtual Experimental Setup}
\label{sec4}
For initial analysis of the networks, we set up a ROS-Gazebo based virtual experimental environment in the OpenAI-Gym framework for robotics \cite{openai} to obtain the training and test data for the network. Figure \ref{ICAPS_fig3a} shows a snapshot of the same. The virtual setup has two turtlebots, Tb1 and Tb2. During the data collection phase, Tb1 remains stationary while tracking and recording the movement of Tb2 using a monocular camera. Tb2 moves in front of Tb1, from point A to point B using a Proportional Integral Derivative (PID) controller that corrects the positional and angular error of the robot. We introduce variation in the PID parameters so that no two trajectories are the same even when Tb2 is moving towards the same goal point. This introduces variance in the local neighbourhood of the trajectories even when the goal point is same. We also recorded video of Tb2 moving towards 4 different target points. These 4 different target points are $(1,0.8)$, $(1.5, -0.8)$, $(2-0.8)$ and $(0.5, -0.5)$ where the position of Tb1 is taken as the origin of the inertial frame. The recorded image frames are converted to gray scale images before being used to train the networks. Altogether we collected about 80 different trajectories (20 trajectories in the local neighbourhood of each of the 4 goal points).
 
\begin{figure*}
	\centering
	\includegraphics[width=0.92\linewidth, keepaspectratio]{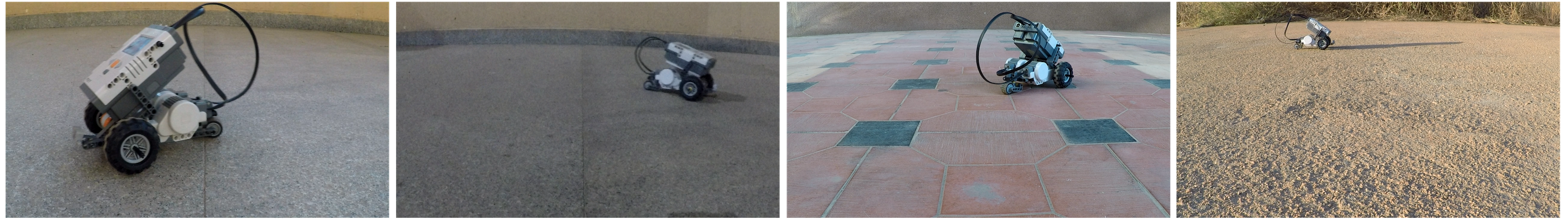}
	
	\caption{The 4 environments from left- Atrium (daylight), Atrium (artificial light), Pavement and Airstrip, respectively}
	\label{ICAPS_fig4}
\end{figure*}

\begin{figure}
	\centering
  \includegraphics[width=.75\linewidth, keepaspectratio]{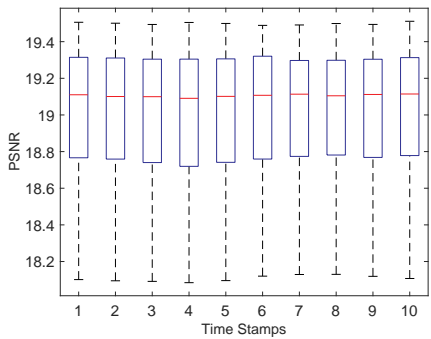}
	\caption{PSNR comparison plot between 2 videos of equal length from two different environment (Atrium daytime and Pavement).}
	\label{ICAPS_fig5}
\end{figure}

\begin{figure*}
        \centering
        \includegraphics[width=.91\textwidth,keepaspectratio]{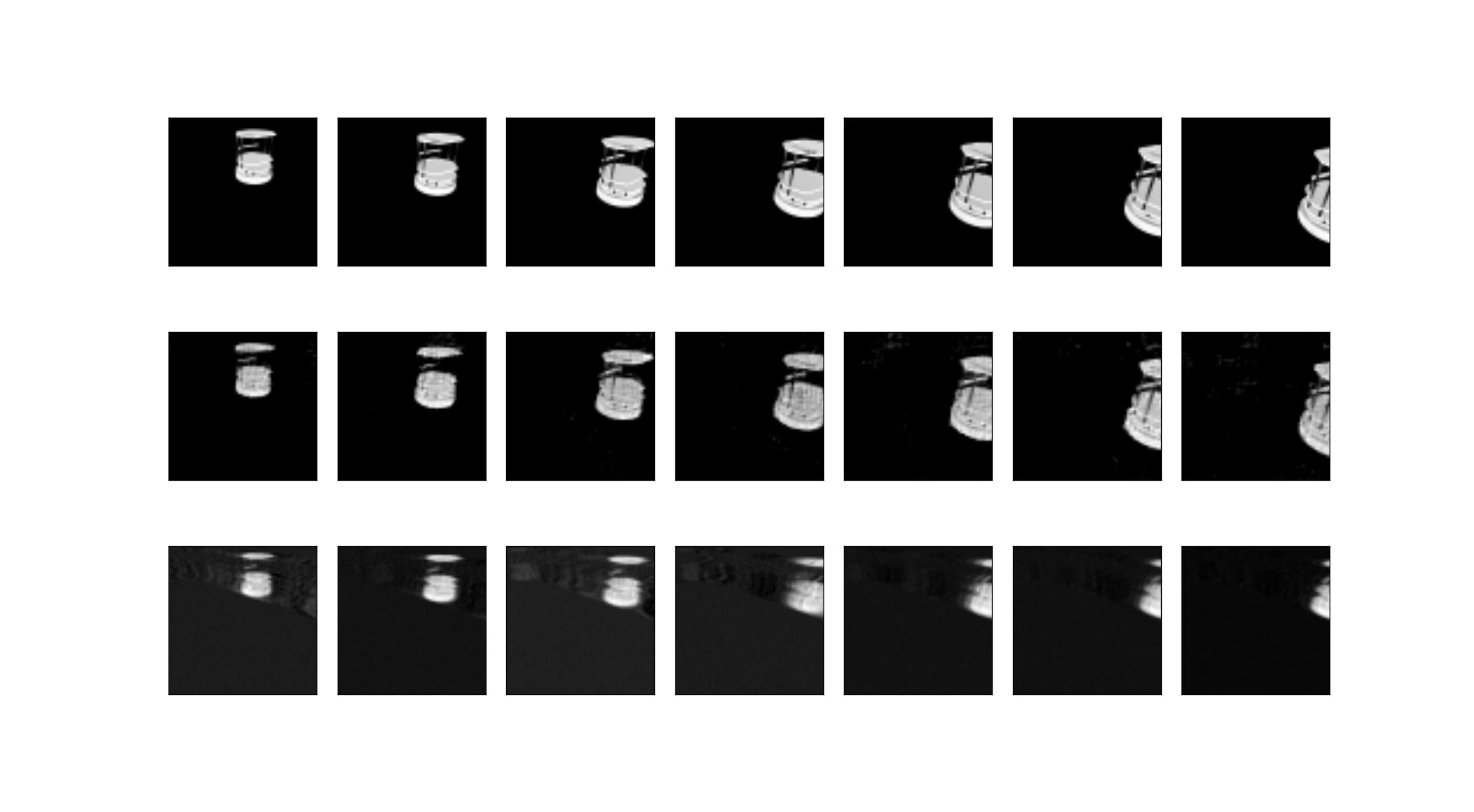}
        \caption{Qualitative comparison on the performance of Fully Connected LSTM network and PROM network on simulated data set. The first row represents the ground truth, second and third row show the estimates by PROM network and the fully connected LSTM network, respectively for time stamps $10$, $15$, $20$, $25$, $30$, $35$ and $40$.}
        \label{ICAPS_fig6}
\end{figure*}

 \begin{figure*}
        \centering
        \includegraphics[width=.7\textwidth,keepaspectratio]{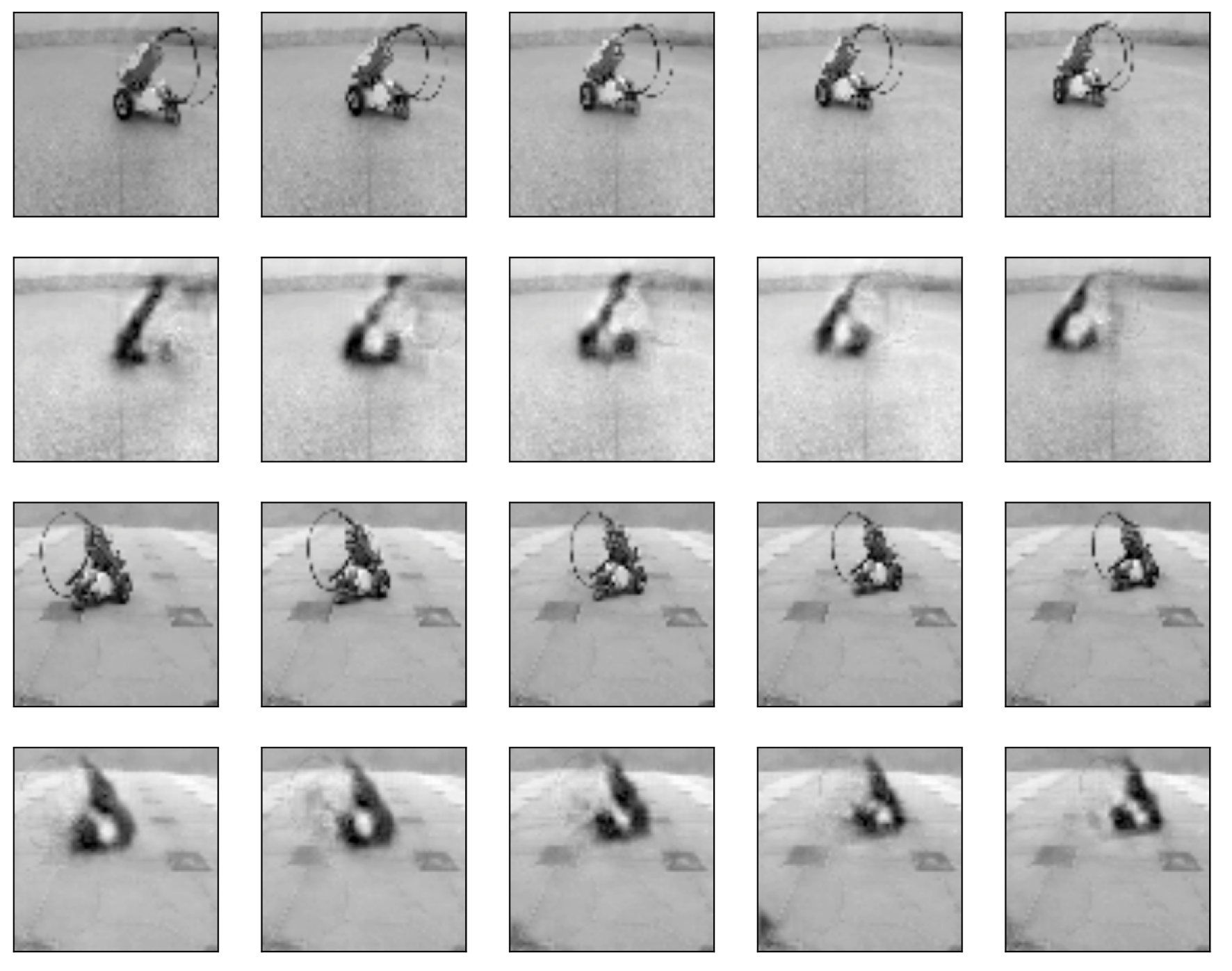}
       
        \caption{Qualitative analysis on the performance of PROM-Net trained on ARM data set. The first and thrid row from top represents the ground truth, second and fourth row show the estimates generated by PROM-Network for time stamps $20$, $30$, $40$, $50$ and $60$. The first 2 rows represent data under artificial lighting conditions and the last 2 rows are from the outdoor environment.}
        \label{ICAPS_fig7}
\end{figure*}
 \section{Real Robot Motion Data-set}
To evaluate the performance of PROM-Net with the real-life data, we created our own Actual Robot Motion (ARM) data-set, using two LEGO Mindstorms under different lighting conditions in 4 different environmental settings- indoor (Atrium) daylight, indoor artificial light, and outdoor (pavement) daylight and outdoor (Airstrip) sunlight. To the best of our knowledge, this is the first of its kind data set where the motion of a mobile robot is captured from the first-person view of another robot. In this section, we present details on the real robot motion data set that was collected from the first person perspective of a LEGO Mindstorms robot (see figure \ref{ICAPS_fig3b}) observing another Mindstorms moving in its field of view.

We recorded the videos using a GoPro Hero 5 Black camera at 30 fps, with a resolution of 720$\times$1280. We later down-sampled it to a resolution of 320$\times$240. For the initial phase of data gathering, we mounted the camera on a LEGO Mindstorms robot to observe the environment and kept it in a stationary state. 
In future, we will add motion to the recording platform to add more versatility to the data which would closely resemble the practical cases seen in a robot path planning problems. The average speed of the moving agent was kept at about 0.665 km/hr (approximately 11 m/s). The recorded videos do not contain any labelled data as they are meant for unsupervised learning algorithms.

 We recorded about 1.5 hours of robot motion of the other LEGO-bot along various trajectories consisting of approximately 120K frames without excluding any particular segments. The GoPro camera offers digital stabilization. We used the narrow-angle shot setting during the recordings. The wide-angle lens of this particular camera produces a significant amount of fish-eye effect for any object moving relatively close to the camera. A wide angle lens will be used in future when we incorporate recordings of unmanned aerial vehicles (UAV) into the data set. Unlike  the autonomous ground vehicles, the high speed operation of UAVs (Average speed of 5m/s) demands long range visual data for effective path planning. Below we describe the various scenarios of the recorded data. The videos are segregated in a 3:1 ratio between training and test data. The data set can be accessed at \url{https://sites.google.com/view/meenakshis/dataset}
 
 \begin{table}
\centering

\begin{tabular}{ |p{1.7cm}||p{1.25cm}|p{1cm}|p{1.2cm}|p{1cm}|  }

 \hline
Types of trajectory & Atrium Daylight & Atrium Night & Pavement & Airstrip\\
 \hline
St. Line   & 4   &4& 4&4\\
Arc   & 4   &4& 4&4\\
Incline L-R  & 4   &4& 4&4\\
Incline R-L  & 4   &4& 4&4\\

 \hline

\end{tabular}\caption{Arrangement of no. of videos in the Data set}
\label{table1}
\end{table}

	




\section{Scenarios}
Among the two Lego Mindstorms bots, one was  remote-controlled via a Bluetooth module to execute four different types of trajectories- Straight path, Inclined path (left to right and right to left)  and Arc; each with three different depths (distance from the mounted-camera) in all of the four different environmental settings (Figure \ref{ICAPS_fig4}). The logistics of the recorded videos in each of the environment for each the 4 different types of trajectories are given in table \ref{table1}.
This was done to incorporate  diversity (Figure \ref{ICAPS_fig5} shows the distribution of PSNR between 2 videos of equal length from 2 different environment) in the data set and to facilitate efficient training of deep networks. Each trajectory in a particular setting was repeated twice for redundancy in a single video.
\subsection{Environment 1, 2: Atrium (Daylight and Artificial Light at Night)}
This setting was used for collecting two different sets of recordings. One was during daytime using natural light (Figure \ref{ICAPS_fig4},$1^{st}$ frame from left ) and the other at night using multiple light sources of white halogens (Figure \ref{ICAPS_fig4}, $2^{nd}$ frame from left). The smooth floor of the atrium results in consistent motion  without any jerks. However, the artificially lit night-scene introduces complexity due to multiple shadow formations (different intensities) of the same object.
\subsection{Environment 3: Pavement}
This was recorded in a sun-lit scene with nearby tree canopy (shadows in the backdrop, Figure \ref{ICAPS_fig4}, $3^{rd}$ frame from left) . The ground (lock-tiles) adds intrinsic inconsistency in motion and is bright-colored.
\subsection{Environment 4: Airstrip}
This was recorded in twilight (resulting in, elongated shadows) and the motion was the most jittery here due to coarseness of the asphalt (Figure \ref{ICAPS_fig4}, $4^{th}$ frame from left). Also, there are tiny insects moving in the background which adds a naturally dynamic clutter.
 
 \section{Results and Analysis}
\label{results}
Initially, we trained the network in the simulated environment. In order to maintain uniformity during training, we used the RMSProp optimizer with a batch size of 64 and learning rate 0.001 for all the networks. Our initial investigation with the simulated data set revealed that even though fully connected LSTM networks (\cite{srivastava}) generates moderately accurate predictions for trajectories in the close neighborhood of the ones it has been trained on, it fails to generalize the robot motion when the test trajectories are unlike any training data it has seen before. The same can be inferred from figure \ref{ICAPS_fig6}. Figure \ref{ICAPS_fig6} also shows that PROM-Net can efficiently approximate the future robot motion for unforeseen test scenarios.

For each of the test cases, we have given the network 10 image frames as input and the network predicted the next 10 frames in future. The reconstructed frames by PROM-Net on the real robot data set for two different environments (indoor with artificial lights and outdoor with sunlight) are shown in figure \ref{ICAPS_fig7}. Even though for this paper we have only presented results with grey-scale images, our network can be very easily modified for RGB inputs.

\begin{figure}[h]
\centering

  \includegraphics[width=0.4\textwidth]{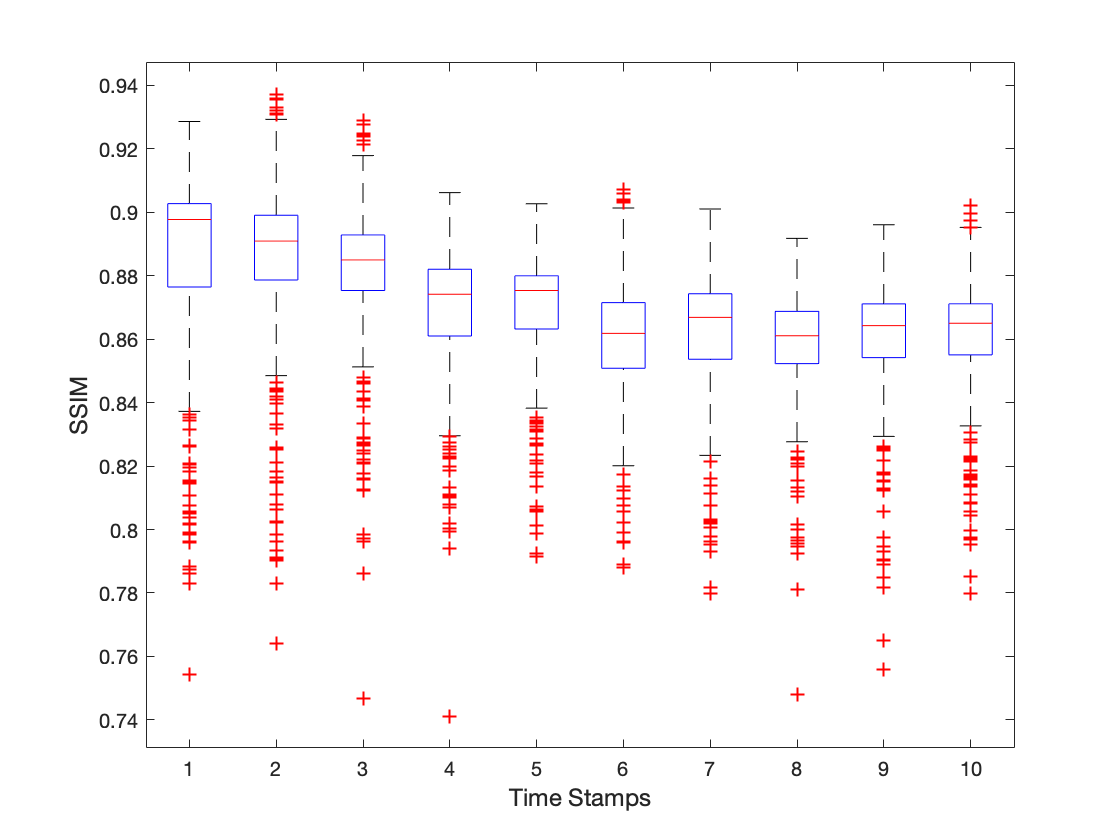}
   
  \caption{SSIM distribution between predicted frames and the ground truth for the 10 time stamps on the ARM data-set.}
  \label{ICAPS_fig8}
 
\end{figure}

\begin{figure}[h]
  \centering
  \includegraphics[width=0.4\textwidth, keepaspectratio]{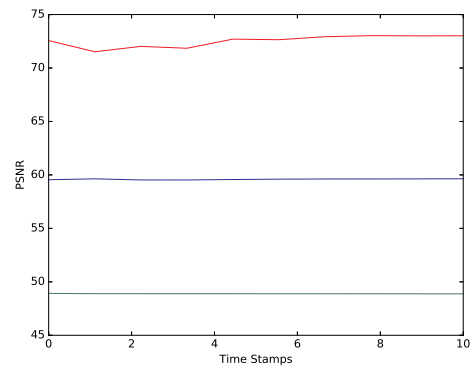}
   
  \caption{PSNR plots for PROM-Net with Real data (red line),  Simulated Data (blue line) and Fully Connected LSTM Network with simulated data (green line).}
 \label{ICAPS_fig9}
\end{figure}

We have given the variation in structural similarity index (SSIM) for all the 10 predicted frames on the real world data set in figure \ref{ICAPS_fig8}. To compare the performance of the proposed prediction network with a FC LSTM network we have given the Peak Signal to Noise Ratio (PSNR) plots for both simulated and real data set in figure \ref{ICAPS_fig9}.  It can be easily inferred from the plots that PROM-Net performs well with both the simulated and real data sets.

From figure \ref{ICAPS_fig7}, we can infer that the blurriness in the predicted frames arises due to the regression losses in convolution layers. As our application is focused on solving path planning problems for robotic agents, we can easily accommodate minor reconstruction loss in the predicted frames. Our intentions are to infer the future direction of motion for the moving objects and PROM-Net has proven to be very effective for that purpose.



\section{Conclusion}
\label{sec7}
We presented a novel light-weight unsupervised learning framework for robot motion prediction problems. A new robot motion data set has been introduced to train and test deep architectures for motion and path planning problems with small scale mobile agents. While the present model is capable of predicting robot motions under stationary condition, a more robust framework is needed in order to estimate future frames where the motion of the robot influences the data collected by the camera sensor. We are already working towards building such models. In  our future work, we plan on designing and testing a vision based MPC on a mobile agent for motion planning in a cluttered dynamic environment. We envisage that reward function would penalize the controller for actions that would move the agent closer to any obstacle and reward it when the area of the obstacle reduces in the predicted frames. We also plan on extending our robot motion data-set with multiple mobile agents (human and robots) moving in the robot workspace.

\end{document}